\documentclass[conference,a4paper]{IEEEtran}
\IEEEoverridecommandlockouts
\usepackage{cite}
\usepackage{amsmath,amssymb,amsfonts}
\usepackage{algorithmic}
\usepackage{graphicx}
\usepackage{textcomp}
\usepackage{xcolor}
\usepackage{color}
\usepackage{subcaption}

\def\BibTeX{{\rm B\kern-.05em{\sc i\kern-.025em b}\kern-.08em
    T\kern-.1667em\lower.7ex\hbox{E}\kern-.125emX}}
\begin{document}

\title{Implicit Neural Representation for Videos \\Based on Residual Connection}

% \author{\IEEEauthorblockN{Taiga Hayami}
% \IEEEauthorblockA{\textit{Graduate School of FSE} \\
% \textit{Waseda University}\\
% Tokyo, Japan \\
% hayatai17@fuji.waseda.jp}
% \and
% \IEEEauthorblockN{Hiroshi Watanabe}
% \IEEEauthorblockA{\textit{Graduate School of FSE} \\
% \textit{Waseda University}\\
% Tokyo, Japan \\
% hiroshi.watanabe@waseda.jp}
% }

\author{\IEEEauthorblockN{Taiga Hayami \qquad Hiroshi Watanabe}
\vspace{2mm}
\IEEEauthorblockA{\textit{Graduate School of Fundamental Science and Engineering, Waseda University} \\
Tokyo, Japan \\
{\small hayatai17@fuji.waseda.jp \qquad hiroshi.watanabe@waseda.jp}}
}

\maketitle

\begin{abstract}

Video compression technology is essential for transmitting and storing videos.
Many video compression methods reduce information in videos by removing high-frequency components and utilizing similarities between frames.
Alternatively, the implicit neural representations (INRs) for videos, which use networks to represent and compress videos through model compression.
A conventional method improves the quality of reconstruction by using frame features.
However, the detailed representation of the frames can be improved.
To improve the quality of reconstructed frames, we propose a method that uses low-resolution frames as residual connection that is considered effective for image reconstruction.
Experimental results show that our method outperforms the existing method, HNeRV, in PSNR for 46 of the 49 videos.

\end{abstract}


\section{Introduction}

Video compression methods reduce the size of the video by removing redundant information while maintaining frame quality.
With the increase in high-resolution video due to advances in technology, the need for high-performance video compression methods is increasing.
Recently, compression methods using implicit neural representations (INRs) have gained attention.
INRs employ neural networks to represent signals like images, scenes, and audio, making it highly versatile for various types of signals.
INRs for videos, such as NeRV \cite{nerv}, represent videos with neural networks.
There are two main types of INRs for videos, index-based and hybrid-based methods.
Index-based methods provide time t as a frame index to the network, while hybrid-based methods provide frame features.
The hybrid-based method, HNeRV \cite{hnerv}, represents a video as a network that generates frames from frame features extracted by an encoder.
However, the fine details of the frame are not adequately represented.

In this paper, we propose implicit neural representation of videos using residual connection.
By employing low-resolution frames as residual connections, the high-frequency components of the frames are made easier to learn.

\section{Related Works}

\subsection{NeRV}

NeRV \cite{nerv} is one of the index-based methods. 
This method represents a video as a network that generates $t$-frame from time $t$, as shown in Fig. \ref{pipeline} (a).
Specifically, a video is represented by a function $f_\theta: \mathbb{R} \to \mathbb{R}^{H\times W\times 3}$ that outputs a frame from frame index $t$.
Video compression is achieved through model compression of the network such as pruning and quantization.

\subsection{HNeRV}

NeRV reconstructs the frames from only frame indexes.
The quality of frame reconstruction is limited by poor input dimensionality and diversity.
To enhance the expressiveness of the input, HNeRV [2] employs frame features instead of frame indexes.
The structure of HNeRV is an auto-encoder, which is a hybrid-based method, as shown in Fig. \ref{pipeline} (b).
An encoder is included to extract frame features.
In HNeRV, the video is represented by the features corresponding to each frame along with a single decoder.
Video compression is achieved by feature compression and model compression of the decoder network.

\section{Proposed Method}

\begin{figure}[tb]
\centerline{\includegraphics[width=1\columnwidth]{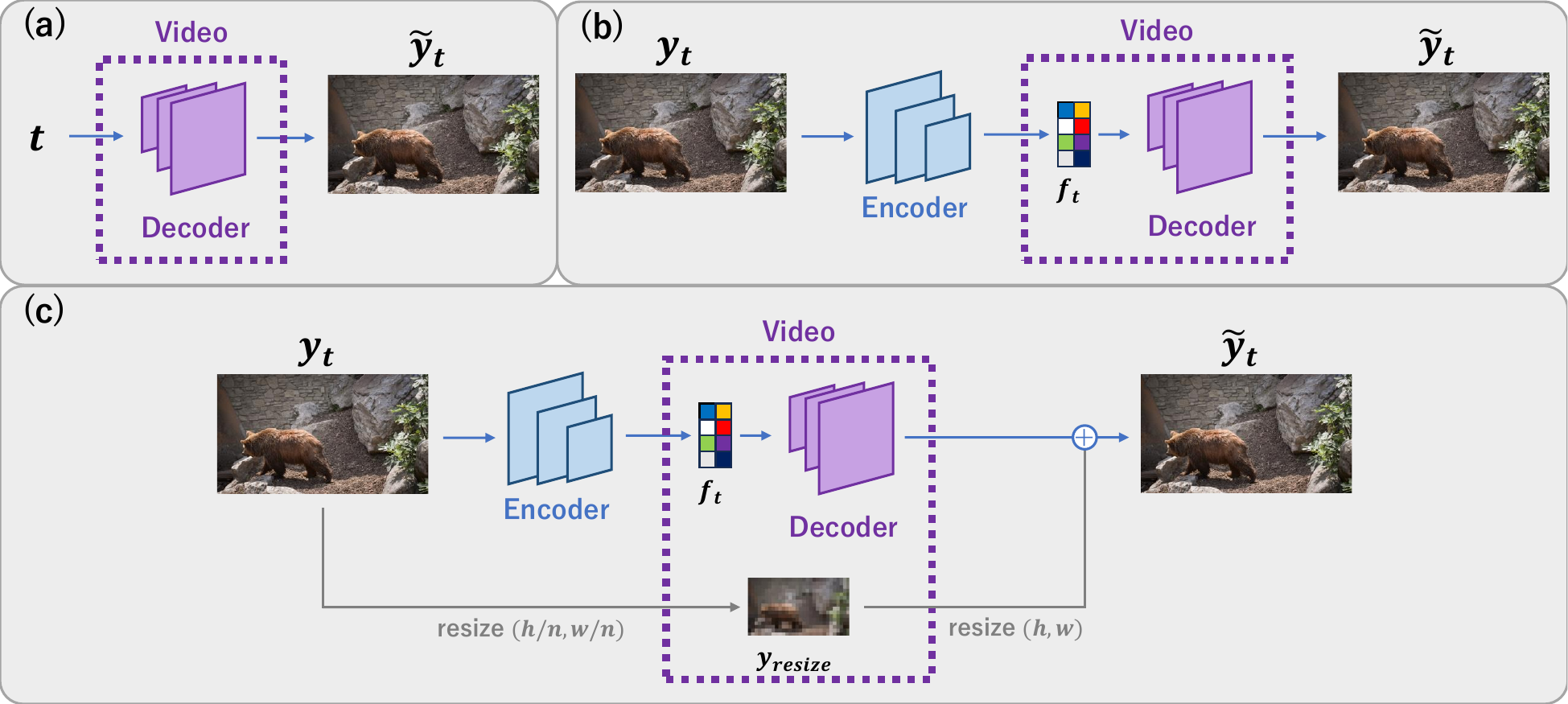}}
\caption{\footnotesize{The pipeline. (a) NeRV. (b) HNeRV. (c) Ours.}}
\label{pipeline}
\end{figure}

\begin{table*}[tb]
\caption{\footnotesize{Comparison of reconstructed videos by PSNR $\uparrow$}}
\label{psnr}
\begin{center}
\begin{tabular}{c|ccccccccccc|c}
\hline
Method & bear         & b-flare      & c-shadow     & d-agility    & elephant     & h-high       & lucia        & motorbike    & rollerblade  & stroller     & train        & ave. \\ \hline \hline
HNeRV  & 26.81        & 29.50        & 33.39        & 32.57        & 30.94        & 31.03        & 29.05        & 31.95        & 35.24        & 32.76        & 30.13        & 31.28 \\ \hline
Ours   &\textbf{27.06}&\textbf{29.59}&\textbf{34.31}&\textbf{33.55}&\textbf{31.23}&\textbf{31.28}&\textbf{29.10}&\textbf{32.93}&\textbf{35.72}&\textbf{33.10}&\textbf{30.30}&\textbf{31.68} \\ \hline
\end{tabular}
\end{center}
\caption{\footnotesize{Comparison of reconstructed videos by MS-SSIM $\uparrow$}}
\label{msssim}
\begin{center}
\begin{tabular}{c|ccccccccccc|c}
\hline
Method & bear          & b-flare       & c-shadow      & d-agility     & elephant      & h-high        & lucia         & motorbike     & rollerblade   & stroller      & train         & ave. \\ \hline \hline
HNeRV  & 0.8632        & 0.8776        & 0.9568        & 0.9562        & 0.9112        &\textbf{0.9341}&\textbf{0.9018}& 0.9305        & 0.9668        & 0.9354        & 0.9255        & 0.9236 \\ \hline
Ours   &\textbf{0.8691}&\textbf{0.8801}&\textbf{0.9624}&\textbf{0.9640}&\textbf{0.9166}& 0.9333        & 0.9008        &\textbf{0.9412}&\textbf{0.9690}&\textbf{0.9394}&\textbf{0.9265}&\textbf{0.9280} \\ \hline
\end{tabular}
\end{center}
% \caption{Comparison of reconstructed videos by MS-SSIM $\uparrow$}
% \label{msssim}
% \begin{center}
% \begin{tabular}{c|ccccccccccc|c}
% \hline
% Method & bear          & b-flare       & c-shadow      & d-agility     & elephant      & h-high        & lucia         & motorbike     & rollerblade   & stroller      & train         & ave. \\ \hline \hline
% HNeRV  & 0.3824        &\textbf{0.3559}& 0.2618        &0.1822         & 0.3553        &\textbf{0.2749}&\textbf{0.2945}& 0.2935        & 0.2079        & 0.3116        &\textbf{0.2972}& 0.2910 \\ \hline
% Ours   &\textbf{0.3772}&0.3577         &\textbf{0.2476}&\textbf{0.1697}&\textbf{0.3454}& 0.2756        &\ 0.2973       &\textbf{0.2766}&\textbf{0.2016}&\textbf{0.3055}& 0.2999        &\textbf{0.2846} \\ \hline
% \end{tabular}
% \end{center}
\end{table*}

In HNeRV, an encoder extracts features from frames.
However, the size of these features is quite small and insufficient for reconstructing the frames.
The decoder that generates frames from these features can capture the basic structure of the video but have difficulty with finer details.
To address this, a residual connection \cite{resnet} to low-resolution frames is used, which helps the decoder focus on detailed representation.
The pipeline of the proposed method is shown in Fig. \ref{pipeline} (c).
Specifically, a low-resolution frame $y_{resize}$ is obtained by reducing the size of frame $y$ with a resizing scale $n$.
The low-resolution frame $y_{resize}$ is resized back to its original size using the bicubic method and the resulting frame $y_{LR}$ is obtained.
The reconstructed frame $\Tilde{y}$ is generated by $y_{LR}$ and the decoder output.
The pipeline is represented as follows,
\begin{equation}\label{diff}
\begin{split}
y_{resize}  &= Resize(y, (y_h/n, y_w/n)), \\
y_{LR} &= Interpolate(y_{resize}, (y_h, y_w)), \\
\Tilde{y} &= y_{LR} + f_\theta(y),
\end{split}
\end{equation}
where $f_{\theta}$ is encoder and decoder.
$y_h$ and $y_w$ are frame height and width.
$T$ frames of video are represented by $T$ features, $T$ low-resolution frames, and a decoder.
The total size of the video representation increases by $y_{resize}$ compared to HNeRV.
When the bit depth of frame is $8$ bits, the increase in bits per pixel (bpp) by $y_{resize}$ is expressed as follows,
\begin{equation}\label{bpp}
bpp_{y_{resize}} = \frac{3\times8}{n\times n}.
\end{equation}
We use feature quantization and model quantization to the decoder parameters as video compression.

\section{Experiment}

\begin{figure}[tb]
\centerline{\includegraphics[width=1\columnwidth]{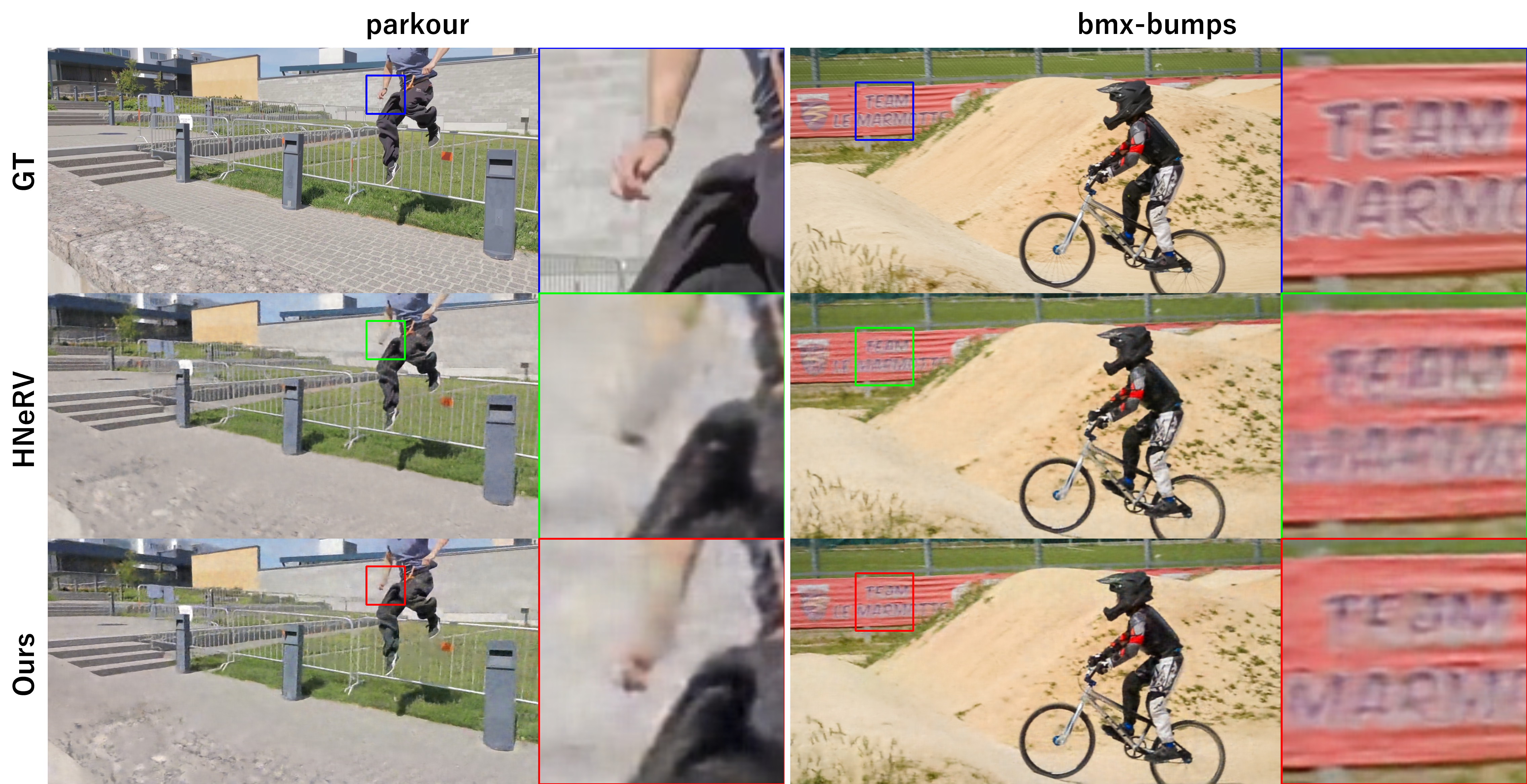}}
\caption{\footnotesize{Visualization of reconstructed videos.}}
\label{result}
\end{figure}

We use videos from the DAVIS \cite{davis} dataset except for the 'motocross-jump' video because HNeRV can not sufficiently represent this video. 
DAVIS dataset has $49$ videos with size $1080\times 1920$, $25$-$104$ frames.
Frame size is cropped to $640\times 1280$.
The resize scale is $n=128$.
Bitrate increase by $1.46\times 10^{-3}$ compared to HNeRV.
The feature quantization factor is $6$ and the model quantization factor is $8$.
The learning rate is $9.9\times 10^{-4}$ for the proposed method and $1\times 10^{-3}$ for HNeRV.
The base model size is $1.5$.
For the evaluation metrics, we used PSNR and MS-SSIM.
%For the evaluation metrics, we used PSNR, MS-SSIM, LPIPS.

\begin{figure}[tb]
\centerline{\includegraphics[width=0.9\columnwidth]{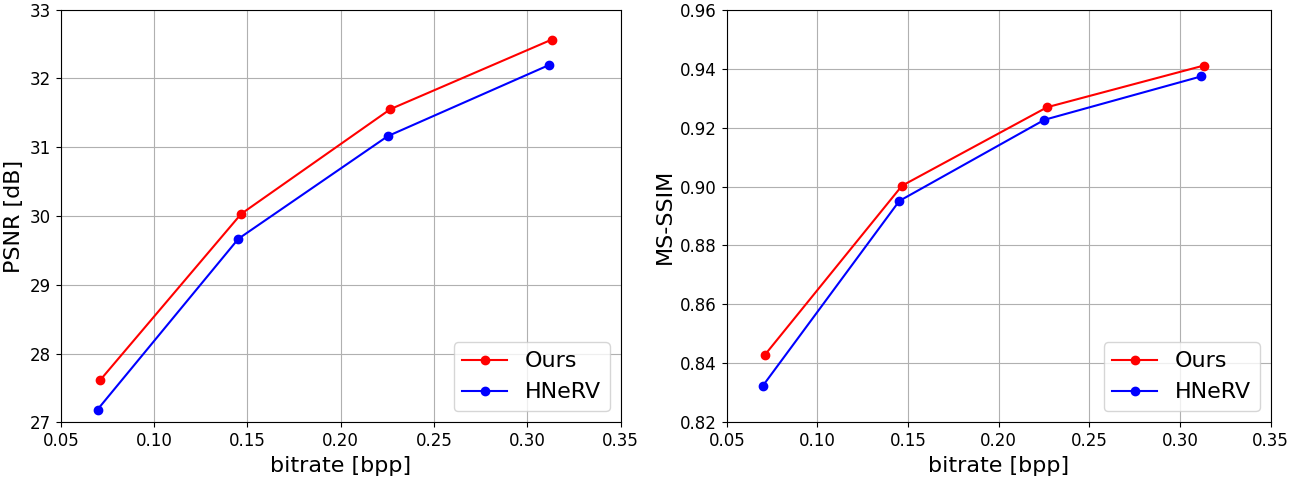}}
\caption{\footnotesize{Video compression results on DAVIS dataset.}}
\label{rd}
\end{figure}

The evaluation results of the video reconstruction for HNeRV and our method are shown in Table \ref{psnr} and \ref{msssim}.
In each table we show results of $11$ videos and the average of $49$ videos.
Our method outperforms HNeRV in PSNR and MS-SSIM for $46$ and $39$ of $49$ videos, respectively.
%Our method outperformed HNeRV in PSNR, MS-SSIM, LPIPS for $46$, $39$, $33$ of $49$ videos, respectively.
Since the network learns residuals, the appropriate resize scale and learning rate need to be selected based on the size of the residuals.
Examples of video reconstruction visualization are shown in Fig. \ref{result}.
In 'parkour' video, the arm in the frame is blurred in HNeRV, while the proposed method is able to represent it.
In 'bmx-bumps' video, the text in the frame is clearly represented.
The results of the video compression are shown in Fig. \ref{rd}.
We change the model size to $0.5$, $1.0$, $1.5$, $2.0$.
The proposed method can achieve improved quality with a small additional bit rate.

\section{Conclusion}

In this paper, we propose a novel implicit neural representation for video using residual connection. 
By adding low-resolution frames as residual connection to the network, detailed expression can be improved.
Experimental results show that our method exceeded HNeRV. 
In the future, we need to consider how to properly select resize scale and learning rate.

\vspace{12pt}
\end{document}